\documentclass[a4paper, 10pt, conference]{ieeeconf}      

\IEEEoverridecommandlockouts                              

\usepackage{amsmath, lipsum} 
\usepackage{amssymb}  
\usepackage{graphicx}
\usepackage{hyperref}
\usepackage{cleveref}
\usepackage{cite}[compress]
\usepackage{graphicx}
\usepackage{cleveref}

\usepackage{breqn}
\graphicspath{ {./figures/} }
\usepackage{float}
\usepackage{flushend} 

\title{\LARGE \bf
Autonomous Parking by Successive Convexification and Compound State Triggers
}

\author{Ali Boyali and Simon Thompson
\thanks{The Authors are with Tier IV, Inc., Tokyo, Japan
        {\tt\small \{ali.boyali, simon.thompson\}@tier4.jp}}%
 }

\begin{document}

\maketitle
\thispagestyle{empty}
\pagestyle{empty}

\begin{abstract}

In this paper, we propose an algorithm for optimal generation of nonholonomic paths for planning parking maneuvers with a kinematic car model. We demonstrate the use of Successive Convexification algorithms (SCvx), which guarantee path feasibility and constraint satisfaction, for parking scenarios. In addition, we formulate obstacle avoidance with state-triggered constraints which enables the use of logical constraints in a continuous formulation of optimization problems. This paper contributes to the optimal nonholonomic path planning literature by demonstrating the use of SCvx and state-triggered constraints which allows the formulation of the parking problem as a single optimisation problem. The resulting  algorithm can be used to plan constrained paths with cusp points in narrow parking environments. 
 
\end{abstract}

\section{INTRODUCTION}
\label{sec:intro}
Motion planning is an essential research domain for autonomous driving and robotic applications. Typically autonomous systems require point-to-point planning between two positions in configuration space. As well as obstacle and path boundary constraints, the  motion of some systems is constrained by their physical design. Motion in nonholonomic systems is constrained by their tangent space, which are characterized by velocity constraints and non-integrable distributions \cite{bloch2005nonholonomic, bloch2003nonholonomic, lavalle2006planning}. Unlike regular constraints on the system state and controls, velocity constraints limit the space of differential motion \cite{latombe2012robot}.

These systems are encountered in the modeling of physical systems in which motion depends linearly on the control \cite{jean2014control} and the number of controls is less than the number of degrees of freedom. Motion planning for such nonholonomic systems is a difficult problem for which no general solution exists, let alone for the more complex problem of considering the existence of obstacles in the environment. 

A wealth of algorithms have been developed to overcome these difficulties. An overall view of these algorithms and related terminology can be found in the books by LaValle \cite{lavalle2006planning} and Fr{\'e}d{\'e}ric \cite{jean2014control}.  The basic structure of the algorithms is to first compute a global solution avoiding all obstacles without considering the feasibility of motion, and then to refine the local behavior of generated path to satisfy the motion constraints \cite{jean2014control, lavalle2006planning}. This two stage process can lead to paths which are not globally optimal.


The autonomous parking problem is an example of nonholonomic motion planning with additional state and control constraints. Motion planning for parking maneuvers also has to consider the presence of cusp points, where the direction of motion of the vehicle changes from forwards to backwards and vice versa. Parking maneuvers can also be tightly constrained by narrow parking spots and neighbouring parked vehicles. In this paper, we present an optimal control based motion planning algorithm for autonomous vehicles. We use the nonholonomic single track kinematic vehicle model in our optimization formulation. Recent developments in convex optimization methods for various nonlinear optimization problems allow us to formulate nonholonomic motion planning in a constrained environment with guaranteed feasibility and recursive constraint satisfaction. 

In this paper, we make the following contribution in the nonholonomic motion planning literature:

\begin{itemize}
    \item We employ successive convexification methods for point-to-point generation of admissible trajectories for a nonholonomic system under tight constraint conditions.

    \item We use state-triggered constraint formulation which allows embedding logical constraints in the continuous formulation of optimal control problems.

    \item We present an algorithm for parking maneuver motion planning using the formulated optimal control problem by which the resultant trajectories resemble Reeds and Shepp (RS) curves \cite{reeds1990optimal}.
\end{itemize}

The rest of the paper is organized as follows. We give a brief summary of the related literature in Section~\ref{sec:nhmp}. A description of the successive convexification and state trigger methods for optimization is given in Section~\ref{sec:scvx}. Section~\ref{sec:vm_parking} details the vehicle model and formulates parking planning as an optimal trajectory generation problem. Simulated results of the proposed system for tightly constrained simulated reverse parking and parallel parking scenarios are presented in Section~\ref{sec:sim}. Finally, in Section~\ref{sec:conclusions} we make concluding remarks and propose future works.

\section{Nonholonomic Motion Planning for Car-Like Systems}
\label{sec:nhmp}
Finding the shortest path between two points for car-like systems was originally studied by Dubins in \cite{dubins1957curves}. The proposed paths only allow forward motion as cusp point are not considered in Dubins's model. Backward motion was introduced three decades later in Reeds and Shepp's work \cite{reeds1990optimal}. In these pioneering works, the shortest-length curve is obtained from a set of curves that are produced by combinations of curve segments. These early nonholonomic path planning methods are geometric methods and generate paths for obstacle-free environments. 

Nonholonomic motion planning problems can be handled well in differential geometric settings analytically as they involve topological structure. Among the analytical solutions to nonholonomic planning is Lafferriere and Sussmann's work \cite{lafferriere1993differential}. In this study, a solution to point-to-point motion planning is derived for driftless control-affine nonholonomic systems of equations \eqref{eq:affine_system}.

\begin{equation}
    \dot{x} = \sum_{i=1}^{n} f_{i}(x)u_{i}
    \label{eq:affine_system}
\end{equation}

The premise of their method is based on obtaining the Philip Hall basis and computing the control variables for an extended system by a formal algebra. The resulting set of differential equations are solved by successive integrations and algebraic operations. If the system of the equations has a nilpotent structure, the resultant solution is exact, otherwise the solution is approximate. Another fundamental study, developed by Murray and Sastry \cite{murray1993steering}, is sinusoidal steering where integrally related control frequencies that steer some of the state components are found, while making sure the rest are kept unchanged at the end of the steering maneuver. 
 
These were followed by the studies that account for obstacle constraints in the same spirit: separating the global and local path planning tasks. Laumond et al. in \cite{laumond1994motion} obtain a global trajectory by concatenating the sub-goals in which the minimum-length curves are produced for any two configurations. Lamiraux et. al in \cite{lamiraux2001smooth}, following the same separation strategy, propose a local planner that approximates local curves with continuous curvature. 

The more recent studies in point-to-point trajectory generation, in particular for autonomous parking, proposals fall under four categories: optimization-based, geometric curve-fitting, machine learning, and sampling-based approaches. For brevity, we review only the optimization methods here. Detailed lists of other approaches can be found in \cite{li2016time, chai2018two, shi2019bilevel, zeng2019unified}.    

In \cite{zhang2018autonomous} and \cite{zeng2019unified}, the authors make use of dual representation of obstacle constraints \cite{boyd2004convex} and solve an identical problem structure by convex optimization methods and nonlinear solvers. While in \cite{zhang2018autonomous} the trajectories are initialized by the Hybrid-A* algorithm, no initialization method is reported in \cite{zeng2019unified}. In both works, the problem is solved for time-optimal trajectories where the boundary positions are enforced in the constraint equations. The distance between two points is not less than 10 [m] in \cite{zeng2019unified}.

Nonlinear Programming (NLP) methods are used to generate time-optimal trajectories for a parallel parking scenario in \cite{li2016time}. The initial trajectories computed off-line are used to facilitate the NLP solution. Another approach seen under optimization-based motion planning is multi-step optimization methods. In \cite{chai2018two}, the trajectories are initialized by a particle swarm optimization method. In the second stage, the trajectories are refined by sequential quadratic programming. The authors propose a bi-level optimal motion planning \cite{shi2019bilevel} in which a first level constraint optimization problem is defined within the constraint equations of the second level optimization problem. The problem is solved by an NLP solver in this study. 

The optimization-based parking proposals differ in computational complexity, obstacle handling and initialization of trajectories. In this study, we elaborate on the use of SCvx algorithms proposed by Mao et al. in \cite{mao2016successive, mao2017successive} to generate nonholonomic vehicle trajectories that may have several cusps in their compositions. The proposed solution is novel in comparison to previous studies in a few ways. First, in this proposal, there is no need to combine other methods such as heuristics, sampling or graph search algorithms to initialize trajectories. The SCvx algorithm can be initialized anywhere in the state space eliminating the need to find feasible initial trajectories steps commonly seen in the literature. Second, we incorporate compound state-triggered constraints (STCs) in maneuver planning to express some of the constraint equations. The state-triggered constraints proposed by Szmuk et al. in \cite{szmuk2018successive, szmuk2019successive} are used to express Boolean decision variables in continuous optimization problems and introduce an additional expansion to the limited family of optimization problems. It also dispenses with the use of constraint equations when they are not active during the solution steps. The constraints are thus activated when only the trigger conditions are active. The last difference is in the structure of the optimization problem in which we force the solver to stitch together separate curve sections by boundary constraints. This strategy yields curves similar to RS-curves. 

\section{Successive Convexification}
\label{sec:scvx}
Solving optimization problems that contain nonlinear differential equations with nonlinear constraints is possible by NLP methods \cite{betts2010practical}. However, there are drawbacks to nonlinear optimization problems rendering them to be impractical for real-time performance. In this case, locally optimal convex problems are more desirable. Convex programs can be solved in polynomial time due to the speed of state-of-the-art convex algorithms and solvers. 

The Successive Convexification (SCvx) algorithm proposed by Mao et al. in \cite{mao2016successive, mao2017successive} is a sequential convex programming method in which nonlinear dynamics and nonlinear constraints are convexified along the previous trajectories obtained from the previous solution. However, convexification and linearization can introduce artificial unboundedness into the problem. This can be remedied by adding a trust region to the problem. Another problem, artificial infeasibility may lead to an early termination of the iterations. Artificial infeasibility is a result of the empty constraint set phenomenon \cite{mao2018successive, reynolds27crawling}. A virtual control term is introduced in the state update equations to ward-off artificial infeasibility. These two algorithmic measures endow the algorithm with the convergence property (existence of convergent sequences) in conjunction with recursive feasibility and recursive constraint satisfaction. 

The literature of SCvx applications has been maturing. For this reason, we will briefly present the formal definition of the algorithm here without dwelling on the individual step details. Such details, implementations, as well as open source code can be found in \cite{reynolds2019dual, szmuk2019successivethesis, malyuta2019discretization, reynolds2020real, code:svengit, code:dmalyuta} (and references therein). The steps for the SCvx algorithms are given in the following subsections.

\subsection{Linearization}

The general structure of a continuous optimization problem for minimization a cost function $J(t_0, t_f, x(t), u(t))$ can be written as:

\begin{align}
    \min_{u(t)}  \quad & J(t_0, t_f, x(t), u(t))  \\ 
    \textrm{s.t.} \quad  & h(x(t), u(t)) =0  \label{eq:equality_constraint}  \\
     & g(x(t), u(t)) \leq 0 \label{eq:inequality_constraint} 
\end{align}

\noindent where the equality and inequality constraints are given by \eqref{eq:equality_constraint} and \eqref{eq:inequality_constraint} as a function of time dependent state $x(t)$ and control $u(t)$ variables. The initial $t_0$ can be fixed, and the final time $t_f$ can be a free variable in the optimization.  In this application, the final time is an unknown. 

Assume that the state differential equations are given as:
\begin{equation*}
\begin{aligned}[c]
    \dot{x(t)} &=f(x(t), u(t), t_f), \quad t_f = \sigma \tau, \\
\end{aligned}
\begin{aligned}[c]
\quad \tau \in [0, \; 1]\\
\end{aligned}
\end{equation*}

\noindent where $\sigma$ is time dilation factor. Following the footsteps in \cite{szmuk2019successivethesis}, the new differential equations assume the form of:

\begin{equation}
    x^{\prime}(\tau)=F(x(\tau), u(\tau), \sigma):=\sigma \cdot f(x(\tau), u(\tau))
    \label{eq:time_dilation}
\end{equation}

The Linear Time Varying (LTV) model of vehicle equations is approximated by the first order Taylor expansion as;
\begin{equation}
\begin{aligned}
&x^{\prime}(\tau) \approx A(\tau) x(\tau)+B(\tau) u(\tau)+S(\tau) \sigma+w(\tau)\\
&A(\tau):=\left.\frac{\partial F}{\partial x}\right|_{\bar{z}(\tau)}\\
&B(\tau):=\left.\frac{\partial F}{\partial u}\right|_{\bar{z}(\tau)}\\
&S(\tau):=\left.\frac{\partial F}{\partial \sigma}\right|_{\bar{z}(\tau)}\\
&w(\tau):=-A(\tau) \bar{x}(\tau)-B(\tau) \bar{u}(\tau)
\end{aligned}
\label{eq:ltv}
\end{equation}

\noindent where the bar notation represents the computed trajectory coordinates obtained by the previous solution, $\bar{z(\tau)}$ at which the linearization takes place with: 
$$\bar{z}(\tau):=\left[\bar{\sigma}\;, \bar{x}^{T}(\tau)\;, \bar{u}^{T}(\tau)\right]^{T} \text { for all } \tau \in [0, 1].$$ This discrete LTV model \eqref{eq:ltv} is expressed in the equality constraint equations \eqref{eq:equality_constraint}. The other nonlinear equations are linearized in a similar manner.

\subsection{Discretization and Scaling}

At each iteration, the continuous equations are discretized and convex sub-problems are solved. The performance of different discretization methods such as Zero or First Order Hold (ZOH, FOH) as well as pseudo-spectral methods are presented in \cite{malyuta2019discretization}. We used FOH in this study. The solution to state differential equation is discretisized using the fundamental matrix solution of the ordinary differential equations. 

In the FOH discretization, the inputs to the model are interpolated between the start and end points of each integration interval. In this case, the discrete equations take the following form \cite{szmuk2019successivethesis, malyuta2019discretization, szmukbehcet2018successive} with the intervals $\tau_k = \frac{k}{K-1}$ and $k \in[0, K]$ for the each step $k$: 

\begin{dmath}
    x^{\prime}(\tau) =A(\tau) x(\tau)+\lambda_{k}^{-}(\tau) B(\tau) u_{\tau}+\lambda_{k}^{+}(\tau) B(\tau) u_{k+1}  + S(\tau) \sigma + w(\tau) 
\end{dmath}

$\forall \tau \in\left[\tau_{k}, \tau_{k+1}\right]$ and the interpolating coefficients:

\begin{align}
    u(\tau) &=\lambda_{k}^{-}(\tau) u_{k}+\lambda_{k}^{+}(\tau) u_{k+1}, & \forall \tau \in\left[\tau_{k}, \tau_{k+1}\right] \nonumber\\
    \lambda_{k}^{-}(\tau)& :=\frac{\tau_{k+1}-\tau}{\tau_{k+1}-\tau_{k}} &  \lambda_{k}^{+}(\tau):=\frac{\tau-\tau_{k}}{\tau_{k+1}-\tau_{k}} \nonumber\\
\end{align}
 
\noindent with the fundamental matrix solution \cite{antsaklis2006linear, hespanha2018linear} to ODEs:

 \begin{equation}
     \Phi_{A}\left(\xi, \tau_{k}\right)=I_{n_{x} \times n_{x}}+\int_{\tau_{k}}^{\xi} A(\zeta) \Phi_{A}\left(\zeta, \tau_{k}\right) d \zeta
 \end{equation}
 
 \noindent where $n_x$ is the state vector dimension in the equations. Using the properties of the fundamental matrix (Theorem II.1 in \cite{malyuta2019discretization}) one can arrive at the non-homogeneous solution  of the Linear Time Varying (LTV) system equations as:
 
  \begin{align}
    x_{k+1} &=A_{k} x_{k}+B_{k}^{-} u_{k} + B_{k}^{+} u_{k+1} + S_{k} \sigma+ w_{k} \\
    A_{k} &:=\Phi_{A}\left(\tau_{k+1}, \tau_{k}\right) \\ 
    B_{k}^{-} &:=A_{k} \int_{\tau_{k}}^{\tau_{k+1}} \Phi_{A}^{-1}\left(\xi, \tau_{k}\right) B(\xi) \lambda_{k}^{-}(\xi) d \xi \\
    B_{k}^{+} &:=A_{k} \int_{\tau_{k}}^{\tau_{k+1}} \Phi_{A}^{-1}\left(\xi, \tau_{k}\right) B(\xi) \lambda_{k}^{+}(\xi) d \xi \\ 
    S_{k} &:=A_{k} \int_{\tau_{k}}^{\tau_{k+1}} \Phi_{A}^{-1}\left(\xi, \tau_{k}\right) S(\xi) d \xi \\
    w_{k} &:=A_{k} \int_{\tau_{k}}^{\tau_{k+1}} \Phi_{A}^{-1}\left(\xi, \tau_{k}\right) w(\xi) d \xi  
 \end{align} 

These matrices are used as the inputs to the solvers to compute the dynamical constraints in the implementation. 

One more subtlety in the numerical optimization is scaling. In general, the states in the equations do not have a similar range of magnitude. This may introduce inconsistencies into the numerical solution. To prevent such problems during the optimization, it is a common practice to normalize the states. One way to do scaling is with a linear transformation. We applied the following transformation to all of the  states $x$ and the inputs $u$ \cite{reynolds2019dual, gill1981practical, ross2018scaling};  

\begin{gather}
    x =D_{\hat{x}} \hat{x}  + C_{\hat{x}} \label{eq:scalingx}\\
    u =D_{\hat{u}} \hat{u}  + C_{\hat{u}} \label{eq:scalingu}\\
    \nonumber
\end{gather}

\noindent where $\hat{x}$ and $\hat{u}$ are the normalized state and control variables. The solver seeks a solution in the normalized variables denoted by the hat notations in \eqref{eq:scalingx}, \eqref{eq:scalingu}. The scaling coefficient matrices $D_{\hat{x}},\; D_{\hat{u}} $ and the centering vectors $C_{\hat{x}}, \; C_{\hat{u}}$ can be computed from the maximum and minimum boundary values of the variables. 

\subsection{Virtual Force for Artificial Infeasibility and Trust Region for Artificial Unboundedness}
A virtual force vector $\nu_k \in \mathcal{R}^{n_x}$ is added to the system equations as an input to prevent artificial infeasibility, making the states one-step reachable: 

\begin{equation}
    x_{k+1}=A_{k} x_{k}+B_{k}^{-} u_{k}+B_{k}^{+} u_{k+1}+w_{k}+\nu_{k}    
\end{equation}

The $\ell_1$ norm is used in the cost function of the virtual force to promote sparsity \eqref{eq:l1}. A high penalty weight $w_{\nu}$  is assigned in the cost for the virtual force so that the solver uses it only when necessary.  The following cost is added to the optimization objective cost function:

\begin{equation}
    J_{v}(\nu):=w_{\nu} \sum_{k \in \overline{\mathcal{K}}}\left\|\nu_{k}\right\|_{1}
    \label{eq:l1}
\end{equation}

As the linearization step might introduce unboundedness, the search space in the optimization variables $[x_k,\;u_k]$ and the final time $\sigma$ are bounded by a trust region. Two forms of trust region formulation are practised in the literature. One can add a hard trust region constraint into the optimization formulation or a trust region cost can be added to the optimization cost using the soft trust region method \cite{benedikter2019convex, szmuk2019successivethesis}. We used the former one: 

\begin{equation}
     \left\lVert \delta x_{k}  \right\lVert_1 + \left\lVert \delta u_{k}  \right\lVert_1 +  \left\lVert  \delta \sigma  \right\lVert_1 < \rho_{tr} 
\end{equation}

\noindent where $\delta x_{k} := \bar{x}(k) - x_k,\;\delta u_{k} := \bar{u}(k) - u_k, \; \delta \sigma := \bar{\sigma} - \sigma$,  and $\rho_{tr}$ is the trust region radius which is scheduled depending on the accuracy of the linear approximation in the model. The details of the trust region scheduling algorithm are given in \cite{mao2018successive, mao2018tutorial}. 

\subsection{Compound State Triggered Constraints - cSTCs} 
State triggered constraints are introduced by Szmuk et al. in \cite{szmuk2018successive} to formulate keep-out zone constraints for the powered descent guidance problem in the aerospace domain. The formulation elegantly embeds logical constraints in the continuous optimal control formulation dispensing with the use of mixed-integer programming, engineering heuristics or continuous consideration of the constraints when they are not active and unnecessary. The formulation of STCs is similar to Linear Complementary Problems (LCP) \cite{cottle1992linear}.

Improved formulations of the state-triggers are presented in \cite{reynolds2019state} and extended to compound STCs formulation in \cite{szmuk2019successive} where logical conjunctions are used to formulate multiple trigger and constraint conditions. Given a logical (binary) trigger function $g(z)$ of an arbitrary solution vector $z$ and the constraint equation $c(z)$ to be satisfied, the formal definition of a continuous STC for the logical constraint:

$$g(z)<0 \Rightarrow c(z) \leq 0$$

\noindent is given by \cite{reynolds2019state}:

\begin{subequations}
\label{eq:stc}
\begin{align} 
\eta &\ge  0  \label{eq:stc1} \\ 
g(z) +\eta &\ge  0 \label{eq:stc2}\\
\eta c(z) & \le 0 \label{eq:stc3}
\end{align}
\end{subequations}

\noindent where $\eta$ is a non-negative slack variable. However, there may arise ambiguity in the formulation \eqref{eq:stc} which does not yield a unique solution for $\eta$. This set of equations are therefore modified in \cite{reynolds2019state} with a new set of constraint conditions as given in \eqref{eq:mstc}.

\begin{subequations}
\label{eq:mstc}
 \begin{align}
    \label{eq:mstc1}
    0 \leq \eta \perp(g(z)+\eta) & \geq 0 \\
    \label{eq:mstc2}
    \eta c(z) & \leq 0
\end{align}
\end{subequations}

The formulation \eqref{eq:mstc} admits an analytical solution $$\eta^{*}:=-\min (g(\boldsymbol{z}), 0)$$ which yields a new conditional constraint formulation as  $$h(\boldsymbol{z}):=-\min (g(\boldsymbol{z}), 0) \cdot c(\boldsymbol{z}) \leq 0.$$ 

The formulations defined above consider only single trigger, single constraint conditions. Compound state trigger conditions are defined in \cite{szmuk2019successive}. We use a compound state trigger for avoiding obstacles in a parking scenario with two 'OR' trigger conditions which triggers a single constraint equation (see below). We refer the readers to the table in \cite{szmuk2019successive} for the complete compound formulations. 

\section{Vehicle Model for Parking and Optimal Trajectories}
\label{sec:vm_parking}
\begin{figure}[ht]
    \centering
    \includegraphics[width=\linewidth]{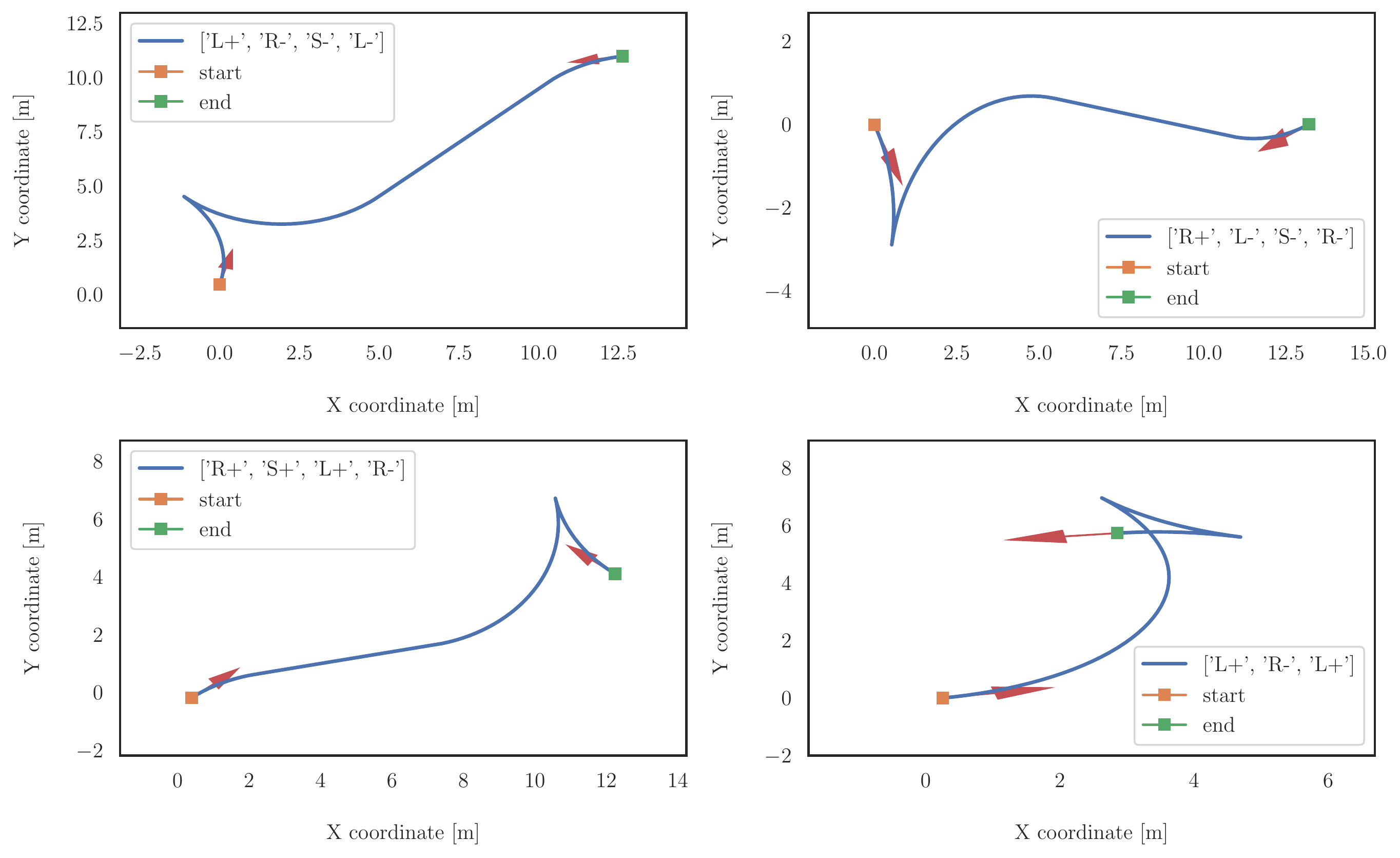}
    \caption{Typical Concatenated RS Curves: L, R, S for Left, Right and Straight, "+" and "-" for Motion Directions.}
    \label{fig:rscurves} 
\end{figure}

Kinematic motion models are suitable for low speed maneuver planning applications, and as such are used in the current study. The classical Dubins and Reeds and Shepp car models are examples of kinematic vehicle models \cite{soueres1998optimal, sussmann1991shortest}, differing in the domain of the control values. The general structure of the kinematic equation is written as:

\begin{equation}
    \left[\begin{array}{c}
{\dot{x_w}} \\
{\dot{y_w}} \\
{\dot{\theta}}
\end{array}\right]=\left[\begin{array}{c}
{\cos \theta} \\
{\sin \theta} \\
{0}
\end{array}\right] u_{1}+\left[\begin{array}{c}
{0} \\
{0} \\
{\frac{1}{R}}
\end{array}\right] u_{2}
\label{eq:motion}
\end{equation}

such that $\left(x_{w}, y_{w}, \theta\right) \in M,(u_1, u_2) \in U$. In this motion model, $x_w, \:y_w$ are the global coordinates, and the controls   $u_1, \:u_2$ are the longitudinal velocity and steering angle respectively. In the Dubins system, the control set is  $U=\{1\} \times\{-1,1\}$ with $R=1$. The car only moves in a forward direction. The Reeds and Shepp car extends \cite{reeds1990optimal, sussmann1991shortest} the Dubins's system with backward motion where the control set is defined as  $U=\{-1, 1\} \times\{-1,1\}$ with $R=1$. In these studies, feasible curves between two points are generated by concatenating at most five motion primitives. In Reeds and Shepp's study, a table of the optimal paths consisting of 48 curve types is given. These paths are represented by motion codes. Four different RS-curves are shown in Fig. \ref{fig:rscurves} where the motion codes are shown. The codes "+" and "-" represents the direction of motion, "L", "S" and "R" represents left turn, straight and right turn motions respectively. The switching points at which steering or direction changes on the curves are computed using the differential geometric approach in  \cite{reeds1990optimal}. 

Sussmann and Tang in \cite{sussmann1991shortest} have investigated the use of RS-curves from an optimal control perspective using the Pontryagin Maximum Principle (PMP) \cite{pontryagin2018mathematical} to characterise the switching points on the curves. The motion model is convexified in \cite{sussmann1991shortest} by setting the control domain as $U=[-1, 1] \times [-1,1]$.  The authors report a reduced number of optimal curves from 48 to 46 by PMP analysis. The computed controls in the study correspond to bang-bang controllers. 

In this paper, we also use a convexified motion model of a RS car. The existence conditions of the optimal curves for the convexified RS car motion are discussed in \cite{sussmann1991shortest, soueres1998optimal}.

\subsection{Application of SCvx to Point-to-Point Curve Generation}
Finding optimal paths between two boundary points in the configuration space without an initial estimate for nonholonomic motion is a hard problem. In the numerical optimal control literature, this issue is handled by proposing different initializing schemes as mentioned in the literature review section along with the analytical solutions proposed by Murray and Sastry \cite{murray1993steering} and Lafferriere and Sussmann \cite{lafferriere1993differential}. The existence of an optimal control for the convexified RS-curve asserts that the control must be convex and that the dynamics of the motion and the cost function must be smooth enough, as defined by Filipov's general theorem for minimum-time problems \cite{soueres1998optimal, sussmann1991shortest}. 

\begin{figure}[ht]
    \centering
    \includegraphics[width=\linewidth]{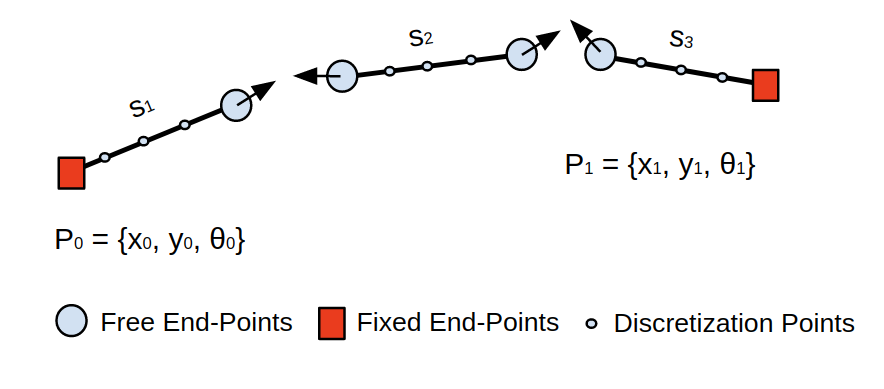}
    \caption{Three separate initial curves, initialized as lines: $s_1$, $s_2$, $s_3$ are the sections. The start $P_0$ and end $P_1$ points are given and fixed.}
    \label{fig:basics} 
\end{figure}

In the early studies and the geometric control theory of nonholonomic systems, solutions for the optimal trajectories are based on the sub-Riemannian geodesic, resulting in a network of locally-connected optimal paths \cite{soueres1998optimal, bloch2003nonholonomic, jean2014control} linking the start and end points in state space. Based on the description of the global path being a network of connected optimal paths, we set up the optimal control problem as finding a few separate continuous curves connecting the state space from the beginning to the end. The start and end points can be arbitrary. In order to keep the computational complexity minimal, and in consideration of typical parking maneuvers, we define three separate curves whose endpoints are required to meet in the state space. This perspective also coincides with the RS or Dubins curve families. Our definition assumes there are at least three separate curves to be connected optimally to achieve the point-to-point path connection in state space. The continuity of the whole curve is enforced in the state space by constraint equations. The basic idea is depicted in the Fig. \ref{fig:basics}. It should be noted that due to the properties of SCvx, the initialization of the connected curve segments is not critical, and in this paper is simply achieved through linear interpolation between the start and end points. The tolerance to the initial solution of the optimization problem is one of the main contributions of this paper.   

In this formulation there are at least three curves, each of which will have their own kinematic motion equation. Defining an augmented state for three curves as $x= [x_1, x_2, x_3]^T $  and augmented control $u = [u_{i1}, \; u_{i2}]$ for $i=1, \;2,\; 3$, the formal definition of the problem can be written in the form of:

\begin{align}
    \min_{u(t_{1:K})} \quad & J(t_0, t_f, x(t), u(t))  \\ 
    \textrm{s.t.} \quad  & h_i(x(t), u(t)) =0   \\
     & g_i(x(t), u(t)) \leq 0  
\end{align}
where the cost is summed over all curve sections, written as:
\begin{align}
    J(t, \hat{x}, \hat{u}) & = J_{\sigma} +  J_{\nu} +  J_{jerk} + J_{length}   \nonumber\\
    J_{\sigma} &= w_{\sigma} \hat{\sigma}   \nonumber \\
    J_{\nu}(\nu) & =w_{\nu} \sum_{i=1}^{3} \left\lVert \nu_{i_{[1:K-1]}}\right\lVert_{1} \nonumber \\
    J_{jerk} &= w_{jerk} \sum_{i=1}^{3} \left[\left\lVert \Delta \hat{u}_{i1_{[1:K-1]}} \right\lVert_2 + \left\lVert \Delta \hat{u}_{i2_{[1:K-1]}} \right\lVert_2  \right]  \nonumber\\
    J_{length} &= w_{length} \sum_{i=1}^{3} \left[\Delta \hat{x}_{{wi}_{[1:K-1]}} + \Delta \hat{y}_{{wi}_{[1:K-1]}} \right] \nonumber \\
    \label{eq:parking_cost}
\end{align}
\begin{equation}
\begin{aligned}[c]
 \mid u \mid    & \leq 1 \\
 u_{11}(K) & = 0 \\
 u_{11}(K) & = u_{21}(0) \\
 u_{21}(K) & = u_{31}(0) \\
 u_{31}(K) & = 0 \\
\end{aligned}
\;
\begin{aligned}[c]
     x_{1}(0) & = P_0 &  \\
     x_{3}(K) & = P_1 &  \\
    x_{1}(K) & = x_{2}(0) &  \\
    x_{2}(K) & = x_{3}(0) &  \\
    \left\lVert \delta \hat{x}_{k}  \right\lVert_1 + \left\lVert \delta \hat{u}_{k}  \right\lVert_1 &+  \left\lVert  \delta \hat{\sigma}  \right\lVert_1  < \rho_{tr} \\
\end{aligned}
\end{equation}


\noindent where the hat notation indicates normalized variables and $w({\cdot})$ refers to the associated variable weight in the optimization. We include the curve length cost in the optimization problem for tuning purposes to find a balance between length minimization and time minimization. One can switch between the two optimization objectives by setting the corresponding weight to zero if required. The jerk cost is added for a smoother control solution. The hard constrained boundary conditions ensure the continuity in the state space. The control for the speed $u_{i1}, \; i \in [1, \; 2, \;3]$ for each curve is also constrained for speed continuity by requesting equal speeds at the curve joints. We do not put boundary constraints on the steering control $u_{i2}$ as it can jump at the cusp joints of the whole curve.

We used norms instead of quadratic forms  whenever it is possible in the implementations as recommended in \cite{cvxadvanced}. The optimization problem can be augmented by the state-triggered constraint for particular problems as we did in the parking simulations described below.

\section{Simulation Of parking maneuvers}
\label{sec:sim}
\begin{figure}[ht]
    \centering
    \includegraphics[width=\linewidth]{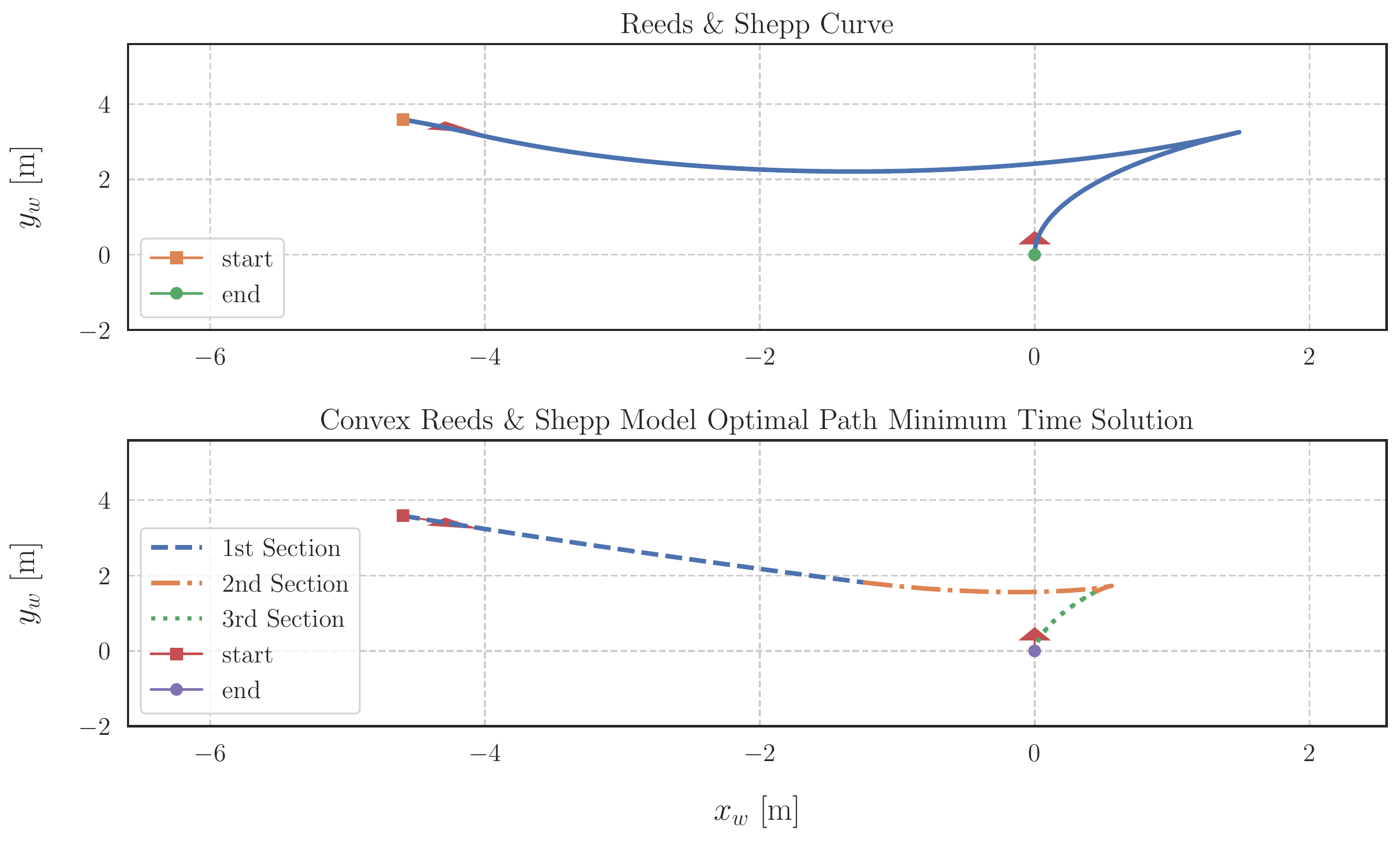}
    \caption{An RS Curve (Top) and A Curve Generated by Optimization (Bottom)}
    \label{fig:rscomp} 
\end{figure}

In simulation, we start with generating unconstrained curves that may have cusps in their sub-sections. A typical unconstrained solution to path generation given a pair of boundary conditions is shown in Fig. \ref{fig:rscomp}. As shown in the figure, the optimization algorithm generates an almost identical shape albeit with some subtle difference. This difference stems from the convex formulation, speed continuity and boundary conditions of the vehicle speed in the SCvx that we enforced.  We set the curvature $\kappa = \frac{1}{R}$ in \eqref{eq:motion} to the maximum curvature value of a turning car. 

In the initial solution, each of the sub trajectories is initialized by interpolating the states between start and end points \cite{szmuk2018successive,szmukbehcet2018successive, reynolds2020real}. The controls are initialized as zero. The control signals can also be initialized exploiting the structure of the solution if it is known a priori. In the shortest time problem for the RS car, the locations and directions of the the control signals have been characterised in \cite{sussmann1991shortest, laumond1994motion} using PMP. However, using these analyses require more work to develop algorithmic initialization procedures. 

\begin{figure}[ht]
    \centering
    \includegraphics[width=\linewidth]{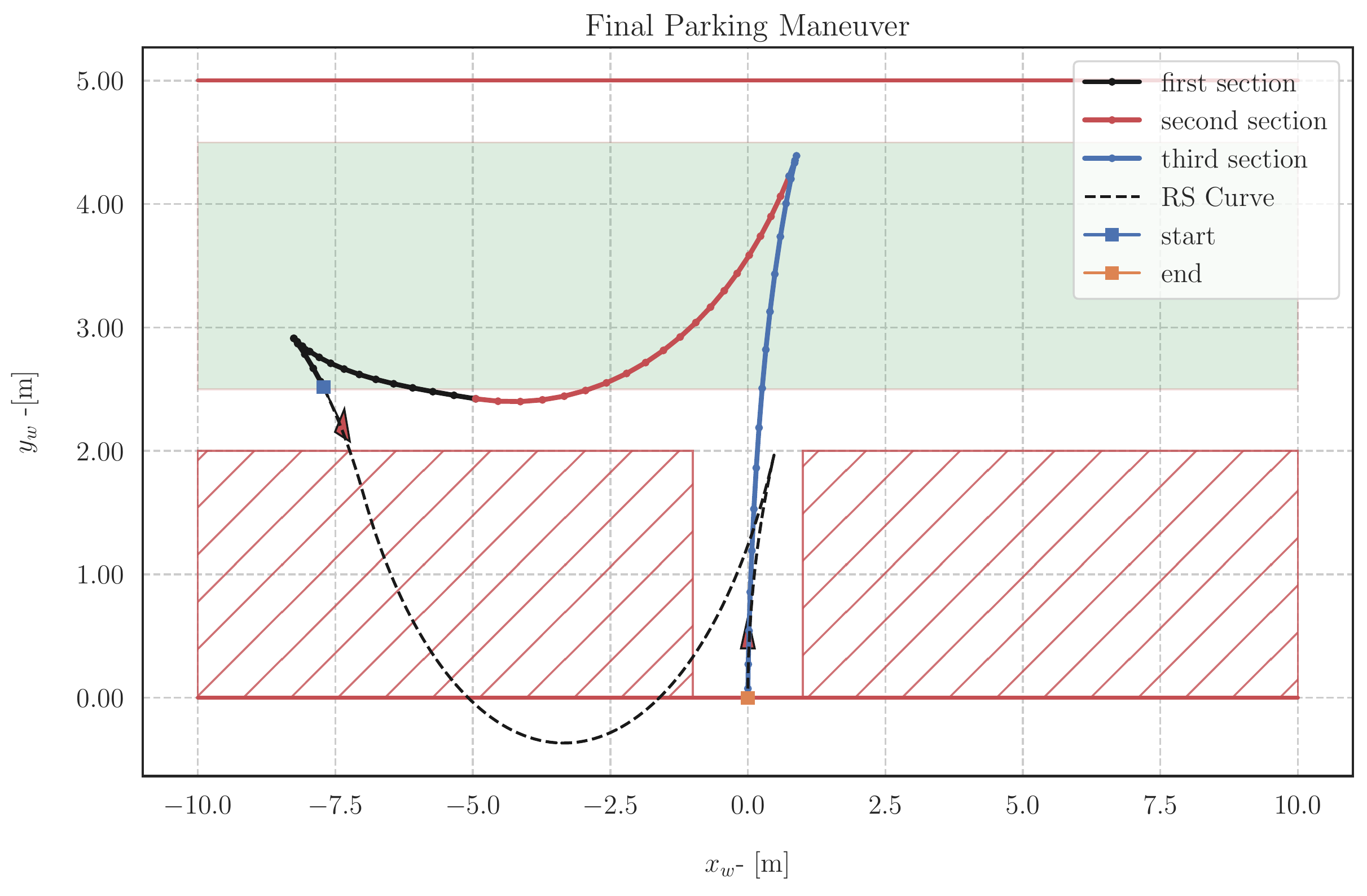}
    \caption{Reverse Parking Scenario}
    \label{fig:reverse} 
\end{figure}

We formulated obstacle avoidance as a compound state trigger constraint for two parking scenarios, reverse and parallel parking. An example result of the reverse parking scenario is shown in Fig. \ref{fig:reverse}. In the reverse parking scenario, the car is required to park in a gap of 2x2 [width x height m] dimensions which is defined by two obstacles (hatched blocks in the figure). The car can start with a random initial position in the shaded area. The heading angle for the initial position is restricted to lie in $[-60, 60]\cup[150, 210]^{\circ}$. 

Only one compound state trigger is formulated in the parking scenario as a minimal demonstration purpose. The trigger function $g(z)$ is a combination of two 'OR' conditions. The compound logical constraint with this trigger function is written as:

$$g(z) = [x_w < -b \quad \textrm{OR} \quad x_w > b] \Rightarrow  c(z) = [y_w \ge 2]$$. 

\noindent where $b=1\; [m]$ is the half-width of parking lot gap between the obstacles. The other constraints in the parking scenario are the upper $y_w \le 5 \: [m]$ and lower bounds $y_w \ge 0 \: [m]$ of the parking location. 

An example RS curve which does not consider the obstacles is shown in the figure (black dashed line) for reference. The trajectory generated by the proposed optimal planner is shown by the solid line (individual curve sections are color coded). As can be seen the optimal curve satisfies all boundary and obstacle constraints, and contains two cusp points, a difficult solution to obtain by classical optimization formulations. 

The control signal in the RS curve algorithm has a bang-bang structure, therefore continuity is not required. The primary aim is to generate nonholonomic paths. If the control signals are to be utilised to control the car, one can enforce equality of the controls signals at the cusp points by equality constraints or adding additional states to the motion equations. In this study, we enforce speed equality as a constraint in the optimization. We added steering equality constraints at the curve joining points only for demonstration of a smooth steering behaviour. However, equality of steering is not always necessary as the steering can jump to different values if the car is in a stopped state. One can define additional state triggers to enforce steering signal equality constraints only when the car is in motion. 

\begin{figure}[ht]
    \centering
    \includegraphics[width=\linewidth]{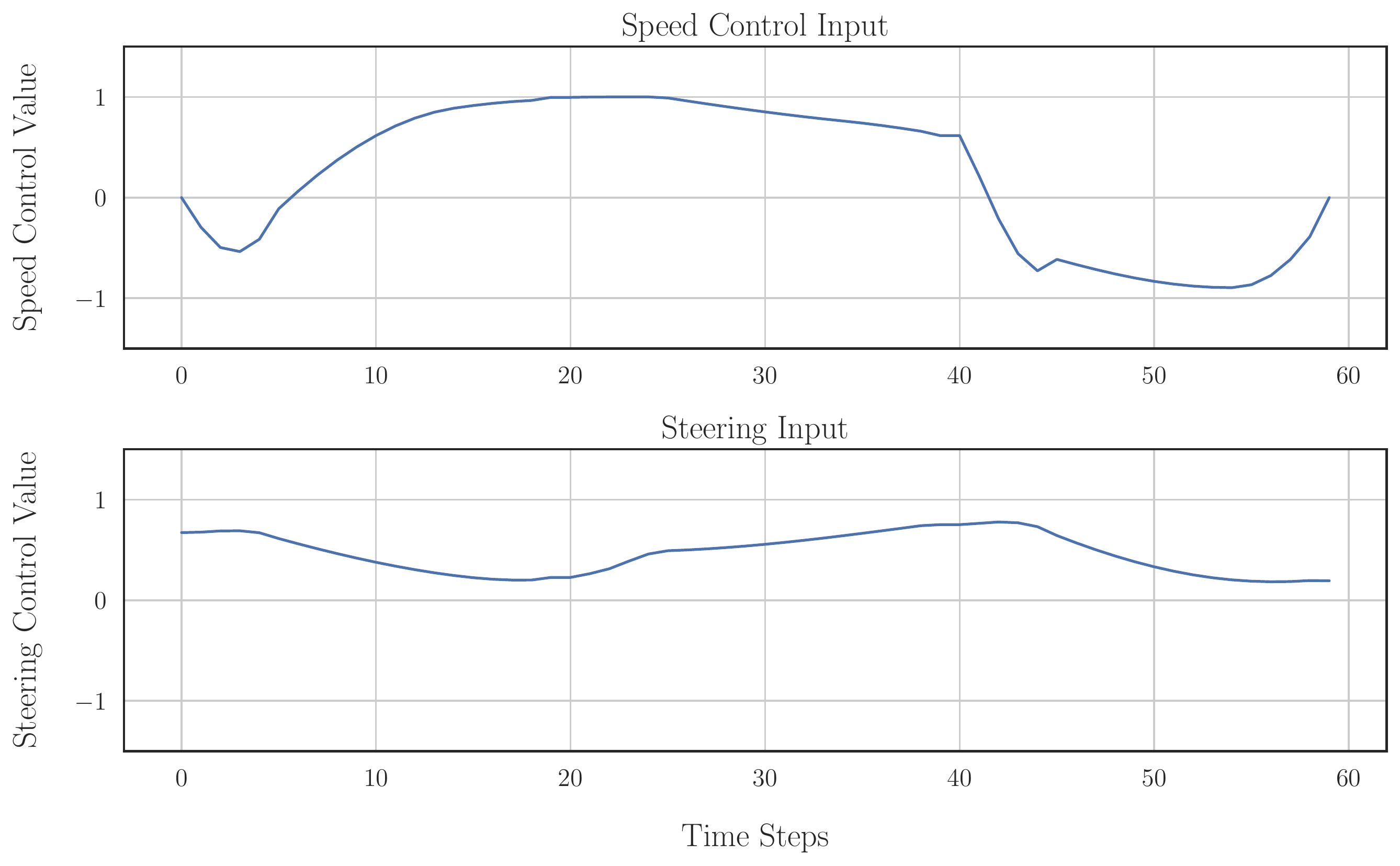}
    \caption{Reverse Parking Scenario - Control Signals}
    \label{fig:reverse_controls} 
\end{figure}

The resulting control signals are shown in Fig. \ref{fig:reverse_controls}. The control solutions are sufficiently smooth and respect the constraints. The forward-backward motion can be seen on the plot of the speed control (sub-figure on the top).

\begin{figure}[ht]
    \centering
    \includegraphics[width=\linewidth]{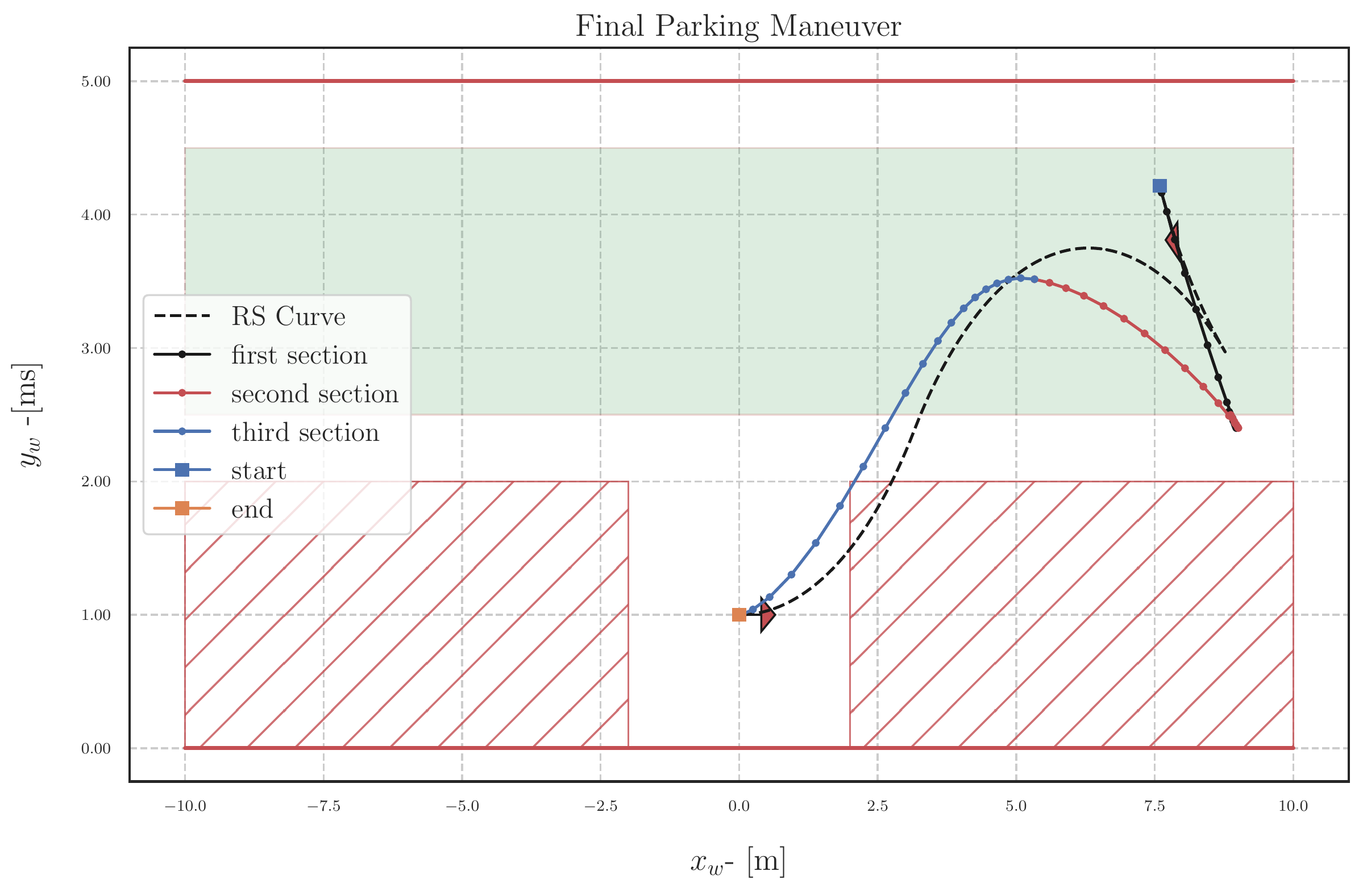}
    \caption{Parallel Parking Scenario}
    \label{fig:parallel} 
\end{figure}

We also simulated the algorithm for a parallel parking scenario in a parking lot, where the parking gap is five meters in width. The RS Curve and optimal solutions with cSTCs are plotted on Fig. \ref{fig:parallel}. 

As seen in the figure, the SCvx algorithm with the state trigger modifies the RS curve counterpart to meet the constraints in the optimization problem and avoids all the obstacles throughout the parking maneuver. 

We implemented the algorithms using Python libraries and a Second Order Conic Solver (SOCP) ECOS \cite{domahidi2013ecos}.  The computation time for the maneuvers varies depending on the number of iterations, number of discretization points and other convergence parameters. We fixed the number of discretization parameters $K=20$ in the simulations. 
 
\section{CONCLUSIONS}
\label{sec:conclusions}
In this paper we proposed a nonholonomic path generation method based on the successive convexification and state-triggered constraint algorithms first introduced for aerospace applications. In the simulation sections, we demonstrated its use for reverse and parallel parking scenarios with minimal implementations. The proposed method can generate feasible parking maneuvers in tightly constrained environments solely through the application of convex optimisation, with out the need for accurate initialization conditions. 

The aim of the paper is to introduce the algorithmic details and implementation by tracking a single point: the center of the rear axle of a car. One can extend the optimization structure by representing a car as a convex polygon and express the state trigger not only a single point but also for the four corner points of the car. Another way of dealing with obstacle avoidance for convex polygon is dual representation of the obstacle constraints as shown in \cite{boyd2004convex, zhang2018autonomous}.  

The solving time varies between 4-8 seconds in our experiments.  The solution in Python might be not convenient for real-time applications, however, C++ implementations of SCvx algorithms are highly promising as reported and demonstrated in \cite{szmuk2019successivethesis, code:svengit}. We also implemented a model predictive control study using SCvx at 60 Hz in C++, and plan a similar implementation for parking maneuver planning, albeit at a lower frequency because of the expanded problem space. The C++ implementation of the proposed path generation and automated parking algorithms will be made available in open source Autoware repository \cite{aw:autoware_repo, kato2018autoware}.











\bibliographystyle{ieeetr}
\bibliography{IEEEexample}

\end{document}